\documentclass[letterpaper, 10 pt, conference]{ieeeconf}  

\usepackage{graphicx}
\usepackage{subcaption}
\usepackage{soul} 
\usepackage{color, xcolor} 
\usepackage{amssymb}
\usepackage{multirow}
\usepackage{colortbl}
\usepackage{color}
\usepackage{graphicx}
\usepackage{subcaption}
\usepackage{amsmath,amsfonts}
\usepackage{booktabs}
\usepackage{bm}

\usepackage{amsmath} 
\usepackage{amssymb}  
\usepackage{color}
\usepackage[linesnumbered,ruled]{algorithm2e}
\usepackage[normalem]{ulem} 
\makeatletter
\let\NAT@parse\undefined
\makeatother
\usepackage{cite}
\usepackage{hyperref}
\hypersetup{pdfstartview=FitH,
            colorlinks=true,
            linkcolor=red,
            anchorcolor=blue,
            citecolor=green
            }

\IEEEoverridecommandlockouts                              

\title{\LARGE \bf
MambaNUT: Nighttime UAV Tracking via Mamba-based Adaptive Curriculum Learning
}

\author{You Wu$^{1}$, Xiangyang Yang$^{1}$, Xucheng Wang$^{2}$, Hengzhou Ye$^{1}$, Dan Zeng$^{3}$, and Shuiwang Li$^{1*}$
\thanks{*Corresponding author}
\thanks{$^{1}$You Wu, Xiangyang Yang, Hengzhou Ye, and Shuiwang Li are with the College of Computer Science and Engineering, Guilin University of Technology, Guilin 541006, China.
\itshape{Email: lishuiwang0721@163.com}}%
\thanks{$^{2}$Xucheng Wang is with the School of Computer Science, Fudan University, Shanghai 200082, China.}
\thanks{$^{3}$Dan Zeng is with the School of Artificial Intelligence, Sun Yat-sen University, Zhuhai 510275, China.}
}

\begin{document}

\maketitle
\thispagestyle{empty}
\pagestyle{empty}

\begin{abstract}

Harnessing low-light enhancement and domain adaptation, nighttime UAV tracking has made substantial strides. However, over-reliance on image enhancement, limited high-quality nighttime data, and a lack of integration between daytime and nighttime trackers hinder the development of an end-to-end trainable framework. Additionally, current ViT-based trackers demand heavy computational resources due to their reliance on the self-attention mechanism.
In this paper, we propose a novel pure Mamba-based tracking framework (MambaNUT) that employs a state space model with linear complexity as its backbone, incorporating a single-stream architecture that integrates feature learning and template-search coupling within Vision Mamba.
We introduce an adaptive curriculum learning (ACL) approach that dynamically adjusts sampling strategies and loss weights, thereby improving the model's ability of generalization.
Our ACL is composed of two levels of curriculum schedulers: (1) sampling scheduler that transforms the data distribution from imbalanced to balanced, as well as from easier (daytime) to harder (nighttime) samples;  (2) loss scheduler that dynamically assigns weights based on the size of the training set and IoU of individual instances.
Exhaustive experiments on multiple  nighttime UAV tracking benchmarks demonstrate that the proposed MambaNUT achieves state-of-the-art performance while requiring lower computational costs.
The code will be available at \url{https://github.com/wuyou3474/MambaNUT}.

\end{abstract}

\section{INTRODUCTION}

Unmanned aerial vehicles (UAV) tracking has emerged as a significant research area in robot vision, with various real-world applications, including navigation \cite{xiao2017uav}, traffic monitoring \cite{tian2011video}, and autonomous landing \cite{gonzalez2021visual}.
While significant advancements utilizing deep neural networks \cite{krizhevsky2012imagenet,he2016deep,dosovitskiy2020image} and large-scale datasets \cite{fan2019lasot,huang2019got,muller2018trackingnet} have led to promising tracking performance in well-illuminated scenarios, existing state-of-the-art (SOTA) UAV trackers \cite{cao2022tctrack,li2023adaptive,lilearning2024} still struggle in more challenging nighttime environments.
Specially, when trackers work under the challenging nighttime conditions, where images captured by UAVs have significantly lower contrast, brightness, and signal-to-noise ratios \cite{ye2022unsupervised} than those captured during the daytime, these approaches often experience a severe degradation in tracking performance.
Therefore, it is essential to develop robust nighttime UAV trackers to enhance the versatility and survivability of UAV vision systems.

\begin{figure}[t]
\centering
\includegraphics[width=0.475\textwidth]{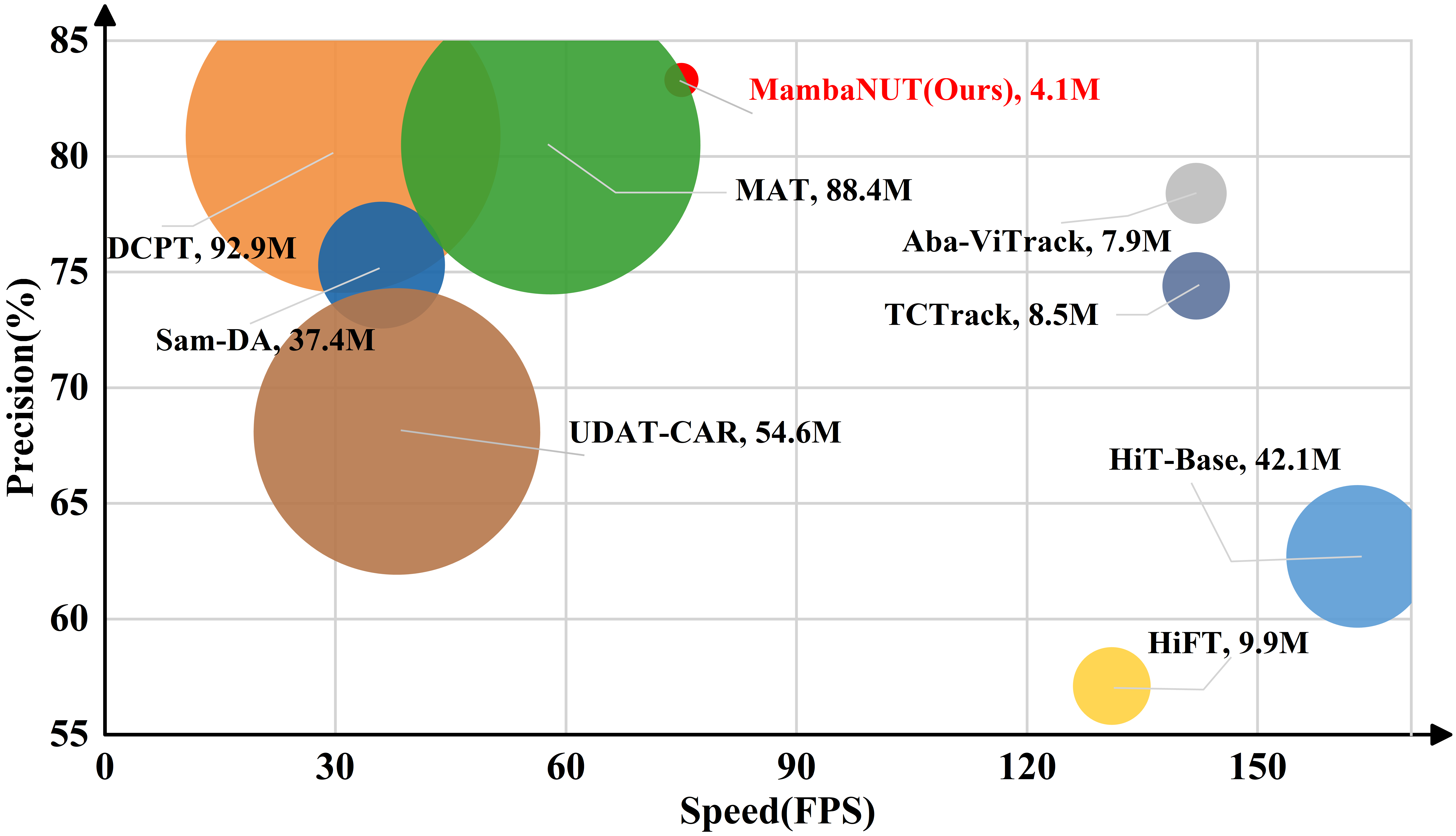}
\caption{Compared to SOTA trackers on NAT2024-1 \cite{fu2024prompt}, our MambaNUT sets a new record with 83.3\% precision and a speed of 75 FPS, while requiring the lowest computational cost.
Note that bubble size reflects the number of parameters, with larger bubbles indicating more parameters.}
\label{fig_Prec_Speed}
\vspace{-20pt}
\end{figure}

Existing nighttime UAV tracking methods can generally be categorized into two types: trackers based on low-light image enhancement techniques and those utilizing domain adaptation (DA).
For the first type of trackers~\cite{ye2021darklighter,ye2022tracker,yao2024enhancing}, a light enhancer is initially used to brighten the nighttime video, followed by a daytime tracker to locate the object.
However, the development of an end-to-end trainable UAV vision system is limited by the reliance on separate enhancement and tracking processes.
The latter trackers use domain adaptation to address the domain discrepancy in nighttime UAV tracking~\cite{ye2022unsupervised,fu2024sam,fu2024prompt}. The DA framework consists of a feature generator and a discriminator, where the generator learns to extract domain-invariant features by deceiving the discriminator using both daytime and nighttime training samples.
Domain adaptation requires large amounts of data for training, but high-quality target domain samples are scarce for nighttime learning.
To address the above issues, DCPT \cite{zhu2024dcpt} introduces an end-to-end trainable architecture that generates darkness clue prompts to enhance the tracking capabilities of a fixed daytime tracker for nighttime operation, enabling robust nighttime UAV tracking without a separate enhancer.
While DCPT demonstrates robust tracking by leveraging critical darkness cues, such ViT-based trackers require significant memory and computational resources because of their reliance on the self-attention mechanism.
Recently, the State Space Model has excelled in modeling long-range dependencies with linear complexity, leading to Mamba's \cite{gu2023mamba} success across visual tasks, particularly in long sequence modeling like video understanding \cite{li2024videomamba} and high-resolution medical image processing \cite{ruan2024vm}.
These successful applications inspired us to adapt Mamba for nighttime UAV tracking, leveraging its long-sequence modeling capabilities to learn robust feature representations in low-illuminated scenarios while maintaining lower computational requirements for efficient nighttime tracking.
Hence, we propose a compact Mamba-based nighttime UAV tracking framework, termed MambaNUT, which adopts a one-stream architecture with a Vision Mamba backbone and a prediction head.

\begin{figure}[t]
\centering
\vspace{-5pt}
\includegraphics[width=0.475\textwidth]{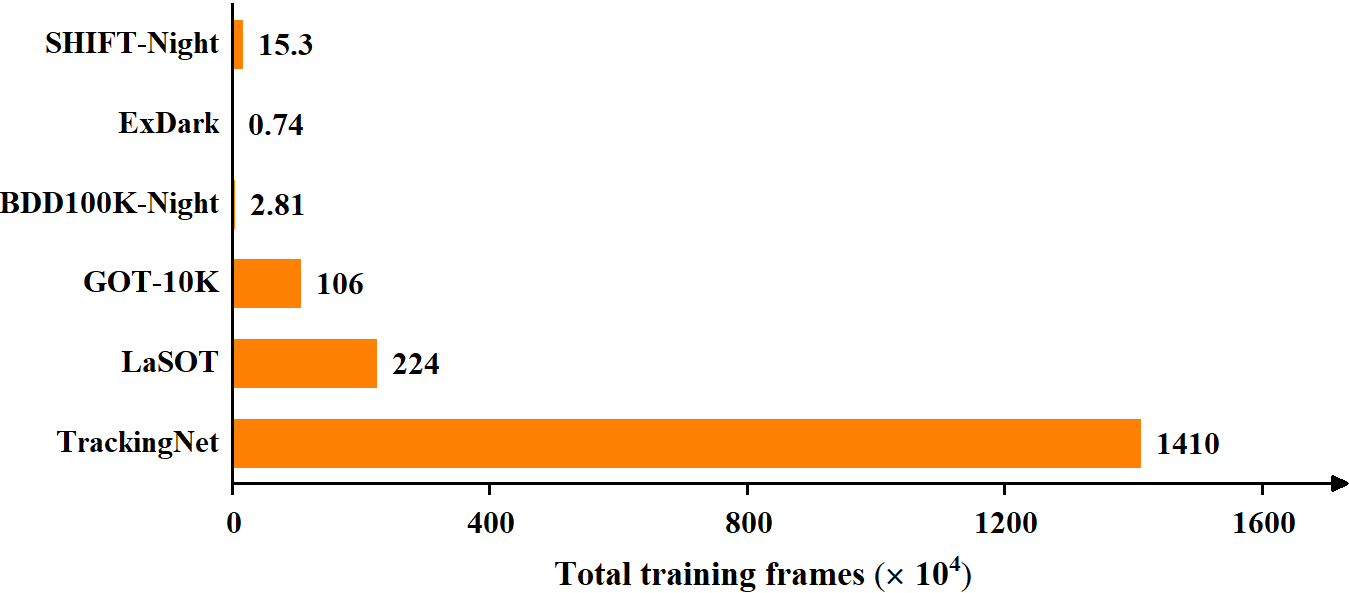}
\caption{Training data distribution varies sharply between daytime and nighttime datasets.}
\label{fig_data_statics}
\vspace{-20pt}
\end{figure}

Additionally, class imbalance is an inherent problem in real-world object detection and classification, often causing algorithms to be biased toward the majority classes~\cite{khan2017cost}.
In visual tracking, there is a similar imbalance in data distribution between day and night, with more data available during the day.
As shown in Fig.~\ref{fig_data_statics}, compared to current large-scale datasets, such as GOT-10K~\cite{huang2019got}, LaSOT~\cite{fan2019lasot}, and TrackingNet~\cite{muller2018trackingnet}, which predominantly consist of daytime images with few or no nighttime images, labeled nighttime data (i.e., SHIFT-Night~\cite{sun2022shift}, ExDark~\cite{loh2019getting}, and BDD100K-Night~\cite{yu2020bdd100k}) remains relatively scarce.
Addressing data imbalance is crucial in this context, as the minority (nighttime) data is the key focus of our work.
Two promising solutions to the imbalanced data learning challenge are resampling~\cite{he2009learning} and cost-sensitive learning~\cite{khan2017cost,charoenphakdee2021classification}.
However, oversampling can lead to overfitting from repeated minority samples, downsampling may discard valuable majority data, and cost-sensitive learning struggles with defining precise costs for samples across different distributions.
Curriculum learning (CL) is the learning paradigm inspired by the way humans and animals learn, gradually progressing from easier to more complex samples during training~\cite{bengio2009curriculum}.
Inspired by CL, we introduce Adaptive Curriculum Learning (ACL) method into our framework to address this issue, based on the following considerations.
We aim for the model to first learn appropriate feature representations during the day to enhance its generalization ability, which will improve the learning of more robust feature representations at night.
Hence, we propose a dynamic sampling strategy that gradually increases the weight of nighttime data and introduce the Adaptive Data Weighted (ADW) loss, which uses a weighting scheme based on training data size and the IoU of individual instances to effectively emphasize hard cases like nighttime data, enhancing calibration performance.
Extensive experiments substantiate the effectiveness of our method and demonstrate that our MambaNUT achieves state-of-the-art performance.
As shown in Fig. \ref{fig_Prec_Speed}, our method sets a new record with a precision of 83.3\%, running efficiently at around 75 frames per second (FPS) on the NAT2024-1 \cite{fu2024prompt} and using only 4.1 million parameters, the lowest in comparison.
The contributions of our work are summarized as follows:

\begin{itemize}

\item We propose a novel Mamba-based tracking framework, termed MambaNUT, which utilizes a purely Mamba-based model for accurate and low-consumption tracking.  To the best of our knowledge, this is the first Mamba-based tracking framework specifically designed for nighttime UAV tracking.

\item We introduce a simple yet effective Adaptive Curriculum Learning strategy to address the imbalance learning between daytime and nighttime data, featuring two curriculum schedulers: a dynamic sampling scheduler and a Adaptive Data Weighted (ADW) loss scheduler.

\item Extensive experiments validate that our MambaNUT surpasses state-of-the-art methods on multiple nighttime tracking benchmarks while requiring lower computational costs.

\end{itemize}

\section{Related work}

\textbf{Nighttime UAV Tracking.}
Real-world UAV tracking applications encounter considerable challenges in low-illumination nighttime scenarios, as generic trackers are primarily designed for daytime conditions.
Recently, low-light enhancement and domain adaptation (DA) have emerged as the two primary methods for improving nighttime UAV tracking performance.
In enhancement-based nighttime UAV tracking \cite{ye2021darklighter,ye2022tracker}, numerous types of enhancers are proposed to improve image illumination prior to processing by the trackers.
However, the limited relationship between low-light image enhancement and UAV tracking leads to suboptimal performance and increased computational costs when enhancers and trackers are integrated in a plug-and-play manner.
For DA training-based nighttime UAV tracking \cite{ye2022unsupervised,fu2024sam,fu2024prompt}, trackers utilize domain adaptation to transfer daytime tracking capabilities to nighttime scenarios.
Unfortunately, DA-based methods incur higher training costs and are limited by the lack of high-quality target domain data for tracking.
Recently, DCPT \cite{zhu2024dcpt} introduced a novel architecture that enables robust nighttime UAV tracking by efficiently generating darkness clue prompts to enhance the tracking capabilities of a fixed daytime tracker for nighttime operation, without the need for a separate enhancer, thus developing an end-to-end trainable vision system.
However, this enhanced tracker burdens resource-limited UAV platforms by adding even more parameters to an already substantial fully ViT-based base tracker, increasing computational resource requirements and hindering efficiency.
In our work, we explore the adaptation of Vision Mamba for nighttime UAV tracking for the first time, leveraging its powerful long-sequence modeling capabilities while ensuring computational costs grow linearly for efficient and accurate tracking.

\textbf{Vision Mamba Models.}
Unlike traditional structured State Space Models \cite{gu2021efficiently}, Mamba employs an input-dependent selection mechanism and a hardware-aware parallel algorithm \cite{gu2023mamba}, enabling it to model long-range dependency linearly with sequence length.
In the field of natural language processing (NLP), it exhibits comparable performance and better efficiency than Transformers in language modeling for long-sequence.
Recently, Mamba's linear complexity in long-range modeling has proven effective and superior across various visual tasks.
In classification tasks, Vim \cite{zhu2024vision} and VMamba \cite{liu2024vmamba} have shown outstanding performance by building on Mamba's success, utilizing a bidirectional scanning mechanism and a four-way scanning mechanism, respectively.
It also exhibits great potential in high-resolution image tasks, with many notable works proposed in medical image segmentation, including VM-UNet \cite{ruan2024vm} and Swin-UMamba \cite{liu2024swin}.
Subsequently, in the field of video, VideoMamba \cite{li2024videomamba} offers a scalable and efficient solution for comprehensive video understanding, encompassing both short-term and long-term content.
MambaTrack \cite{xiao2024mambatrack} explores a Mamba-based learning motion model for multiple object tracking.
In our work,  we propose a novel Mamba-based framework for nighttime UAV tracking that incorporates a Adaptive Curriculum Learning (ACL) method to adaptively optimize the sampling strategy and loss weight, enhancing generalization and discrimination in night tracking.

\textbf{Curriculum learning.}
The concept of curriculum learning (CL), first proposed in \cite{bengio2009curriculum}, shows that the strategy of learning from easy to hard significantly enhances the generalization of deep models.
While these approaches \cite{basu2013teaching,hacohen2019power,khan2011humans} improve convergence speed and local minima quality, pre-determining the order can create inconsistencies between the fixed curriculum and the model being learned.
To address this, Kumar et al. \cite{kumar2010self} proposed the concept of self-paced learning, where the curriculum is constructed dynamically and without supervision to adjust to the learner's pace.
This seminal concept has inspired numerous variations across a range of computer vision applications, including classification ~\cite{wang2019dynamic}, action recognition~\cite{tong2022semi}, and object detection~\cite{zhang2017bridging,soviany2021curriculum}.
Despite its efficacy in these domains, the exploration of curriculum learning in the context of visual tracking remains limited.
In contrast, our work is the first to explore the integration of Vision Mamba with curriculum learning in a unified framework for nighttime UAV tracking, introducing two levels of curriculum schedulers: a dynamic sampling scheduler and a ADW loss scheduler.

\begin{figure*}[t]
\centering
\includegraphics[width=0.95\textwidth]{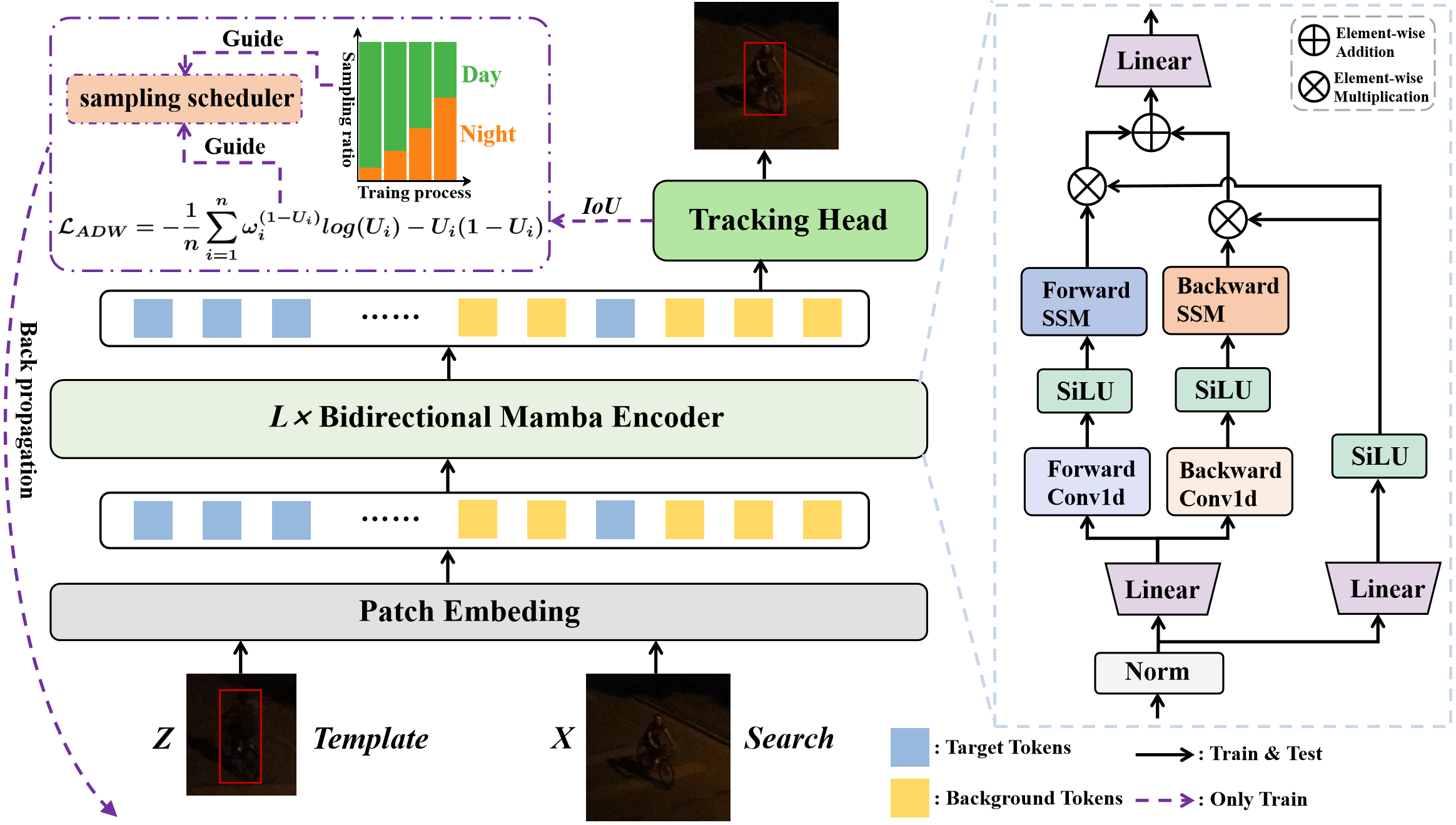}
\caption{Overview of the proposed MambaNUT framework. It includes a Vision Mamba backbone and a tracking head, integrating an adaptive curriculum learning (ACL) approach with two schedulers: (1) a sampling scheduler that balances the data distribution from easier (daytime) to harder (nighttime) samples, and (2) a loss scheduler that assigns weights based on training data size and IoU of individual instances.}
\label{fig_MambaNUT_overview}
\vspace{-5pt}
\end{figure*}

\section{Methodology}

In this section, we detail the proposed end-to-end tracking framework, termed MambaNUT.
First,  we begin with the preliminary of state space models (SSM) and the Mamba.
Then, we introduce the Adaptive Curriculum Learning (ACL) strategy for addressing imbalanced data learning problems, which include  two schedulers for sampling and loss backward propagation.
Last, the overall architecture of the our MambaNUT was described in detail, as shown in Fig. \ref{fig_MambaNUT_overview}.

\subsection{Preliminary}

The raw State Space Model (SSM) is developed for the continuous system, whih is derived from the classical Kalman filter \cite{kalman1960new}.
It maps the 1-dimensional sequence $x(t)\in \mathbb{R}^L\mapsto y(t) \in \mathbb{R}^L$ via a learnable hidden state $h(t) \in \mathbb{R}^N$.
In the continuous state, the specific expression of SSM is formulated by a set of first-order following linear ordinary differential equations:
\begin{equation}
\begin{aligned}
h'(t)&=\mathbf{A} h(t)+\mathbf{B} x(t), \\ 
y(t)&=\mathbf{C} h(t)
\end{aligned}
\end{equation}
where matrices $\mathbf{A} \in \mathbb{R}^{N \times N}$ represents the evolution parameters and $\mathbf{B} \in \mathbb{R}^{N \times 1}$, $\mathbf{C} \in \mathbb{R}^{1 \times N}$ are the projection parameters.

The modern SSMs, i.e., S4 \cite{gu2021efficiently} and Mamba \cite{gu2023mamba} are the discrete forms of this continuous state. By introducing the time scale parameter $\Delta$, the process of discretization is typically accomplished using a rule called zero-order hold (ZOH):

\begin{equation}
\begin{aligned}
\overline{\mathbf{A}}&=\exp (\Delta \mathbf{A}), \\
\overline{\mathbf{B}}&=(\Delta \mathbf{A})^{-1}(\exp (\Delta \mathbf{A})-\mathbf{I}) \cdot \Delta \mathbf{B}, \\
h_{t}&=\overline{\mathbf{A}} h_{t-1}+\overline{\mathbf{B}} x_{t}, \\
y_{t}&=\mathbf{C} h_{t} .
\end{aligned}
\label{eq_ssm}
\end{equation}
where $\overline{\mathbf{A}}$ and $\overline{\mathbf{B}}$ are  the discrete counterparts of parameters $\mathbf{A}$ and $\mathbf{B}$.
$h_{t}$ and $h_{t-1}$ denote the discrete hidden states at various time steps, respectively.
Unlike traditional models that depend heavily on linear time-invariant state space models (SSMs), Mamba \cite{gu2023mamba} improves the SSM by incorporating the Selective Scan Mechanism (S6) as its core operator.
 This is achieved by parameterizing the SSM parameters $\mathbf{B} \in \mathbb{R}^{B \times L \times N}$, $\mathbf{C} \in \mathbb{R}^{B \times L \times N}$ and $\Delta \in \mathbb{R}^{B \times L \times D}$ using linear projection based on the input $x \in \mathbb{R}^{B \times L \times D}$.


\subsection{Overview}

As shown in Fig. \ref{fig_MambaNUT_overview}, our proposed MambaNUT adopts a one-stream framework, which includes a Visison Mamba backbone and a tracking head.
The framework takes a pair of images as input, namely the template image  $Z\in\mathbb{R}^{3\times H_{z}\times W_{z}}$ and the search image $X\in\mathbb{R}^{3\times H_{x}\times W_{x}}$.
These images are respectively split and flattened into patch sequences $P\times P$, and the number of patches for $Z$ and $X$ are $P_{z}=H_{z}\times W_{z}/P^2$ and $P_{x}=H_{x}\times W_{x}/P^2$.
The features extracted from the Vision Mamba backbone are input into the prediction head to generate the final tracking results.
To enhance the learning of robust feature representations from nighttime samples, we propose a Adaptive Curriculum Learning (ACL) strategy for the imbalanced data learning problem, which features two-level curriculum schedulers:
(1) a sampling scheduler that transforms the data distribution from imbalanced to balanced, as well as from easier (daytime) to harder (nighttime) samples; (2) a data-dependent dynamically weighted loss function that assigns weights based on size of training data and the IoU of individual instances. 
The details of the proposed method will be elaborated in the subsequent subsections.

\subsection{Adaptive Curriculum Learning (ACL)}

Sampling is one of the simple and effective methods to deal with imbalanced data learning.
Our sampling scheduler is a key element of the proposed Adaptive Curriculum Learning (ACL) strategy, dynamically adapting the daytime and nighttime data distribution in a batch from imbalanced to balanced throughout the training process.
During training, we assign equal sampling weights to all datasets within each epoch; however, for nighttime datasets, their weights are adjusted by dividing by a constant and multiplying by the training epochs, resulting in a smaller initial proportion of nighttime data that gradually increases as training advances.
Given a training dataset $d$, its assigned sampling weight can be expressed as follows:
\begin{equation}
\begin{split}
w_d=\left\{\begin{array}{ll}
\frac{1}{\theta}*e, & \text { if } d \text { belongs to } \mathcal{N} \\
1, & \text { otherwise }
\end{array}\right.
\end{split}
\label{eq_weight_dataset}
\end{equation}
where $e$ refers to current training epoch, $\theta$ represents a constant, set to 150, which is half of the total training epochs.
$\mathcal{N}$ is the set of all nighttime datasets.
Then, the final sampling ratio for a given dataset among the combination of training sets is: $r_i = w_i/\sum_{i=1}^{N}w_i$, where $N$ denotes the number of training datasets.
Usually, the model learns lots of easy (daytime) samples in early stage of the training process. Going further with the training process, the data distribution between daytime and nighttime is gradually getting balanced. 
During the training phase, the back propagation algorithm updates the network's parameters based on the errors computed by the loss function.
Training the tracking model with equal weights for samples under varying lighting conditions leads to imbalanced adaptation, caused by the significant distribution disparity between daytime and nighttime, where nighttime images have lower contrast, brightness, and signal-to-noise ratio, causing the tracker to be biased toward daytime conditions.
In our work, the minority nighttime samples are the key instances of interest in this learning task.

In view of this, we introduce an Adaptive Data Weighted (ADW) loss that assigns weights based on the size of the training set and IoU, thereby dynamically focusing more on the challenging minority samples, i.e., the nighttime data.
For convenience, let the $IoU$ between the predicted boxes and the ground truth be denoted as $U$.
Thus, $U_i$ is the $IoU$ of the instance $x_i$.
Inspired by \cite{fernando2021dynamically}, the proposed ADW is formulated as follows:
\begin{equation}
\begin{aligned}
\mathcal{L}_{ADW}=-\frac{1}{n}\sum_{i=1}^{n} \omega_i^{(1-U_i)}log(U_i)-U_i(1-U_i)
\end{aligned}
\end{equation}
where $\omega_i$ is a hyperparameter determined by the size of training set.
In the context of classification, $\omega_i$ is typically inversely proportional to the frequency of the classes, allowing it to effectively penalize the majority classes.
In our implementation, we define $\omega_i$ as the logic ratio of the size of the largest training set, given by $\omega_i = \log(N_{max}/N_j) + 0.5$, where $N_{max}$ denotes the total sample size of the largest training dataset (i.e., one of the daytime datasets), and $N_j$ represents the total sample size of the dataset to which the $i$-th sample belongs.
Adding 0.5 to the log weights to avoid situations where the weight equals zero.
If an instance belongs to a dataset with a large number of samples, its weight is relatively small, and vice versa.
With this setup, the minority nighttime data contribute more to the network's gradient calculation, allowing the network to focus less on the majority daytime data and more on the minority during training.
$U_i(1-U_i)$ serves as a regularization term that penalizes uncertain predictions, encouraging the model to produce more confident results. As the modulating factor, $\omega^{(1 - U_i)}$ directs the network to focus more on samples with lower IoU values.

\begin{table*}
\centering
\caption{State-of-the-art comparison on the NAT2024-1 \cite{fu2024prompt}, NAT2021 \cite{ye2022unsupervised}, and UAVDark135 \cite{li2022all} benchmarks. The top two results are highlighted in \textcolor{red}{\textbf{red}} and \textcolor{blue}{\textbf{blue}}, respectively. Note that the percent symbol (\%) is excluded for precision (Prec.), normalized precision (Norm.Prec.), and success rate (Succ.) values.}
\label{table:overall_performance}
\resizebox{7.0in}{!}{
\begin{tabular}{cc|ccc|ccc|ccc|ccc} 
\toprule[1pt]
\multirow{2}{*}{Method} & \multirow{2}{*}{Source} & \multicolumn{3}{c|}{NAT2024-1\cite{fu2024prompt}}                                                                                                    & \multicolumn{3}{c|}{NAT2021\cite{ye2022unsupervised}}                                                                                                      & \multicolumn{3}{c|}{UAVDark135\cite{li2022all}}                                                                                                   & \multirow{2}{*}{Avg.FPS} & \multirow{2}{*}{FLOPs(GMac)} & \multirow{2}{*}{Params.(M)}  \\
                        &                         & Prec.                                      & Norm.Prec.                                  & Succ.                                      & Prec.                                      & Norm.Prec.                                  & Succ.                                      & Prec.                                      & Norm.Prec.                                  & Succ.                                      &                          &                           &                             \\ 
\hline
\textbf{MambaNUT}          & \textbf{Ours}   & \textcolor{red}{\textbf{83.3}}            & \textcolor{red}{\textbf{76.9}}            & \textcolor{red}{\textbf{63.6}}            & \textcolor{red}{\textbf{70.1}}            & \textcolor{red}{\textbf{64.6}}            & \textcolor{blue}{\textbf{52.4}}           & \textcolor{red}{\textbf{70.0}}            & \textcolor{blue}{\textbf{69.3}}           & \textcolor{red}{\textbf{57.1}}            & 72                       & \textbf{1.1}                     & \textbf{4.1}                       \\
\hline
DCPT \cite{zhu2024dcpt}                    & ICRA 24                 & \textcolor{blue}{\textbf{80.9}}           & \textcolor{blue}{\textbf{75.4}} & \textcolor{blue}{\textbf{62.1}}           & \textcolor{blue}{\textbf{69.0}}           & \textcolor{blue}{\textbf{63.5}}           & \textcolor{red}{\textbf{52.6}}                                      & \textcolor{blue}{\textbf{69.2}}           & \textcolor{red}{\textbf{69.8}}                                      & \textcolor{blue}{\textbf{56.7}}           & 35                       & 29.4                    & 92.9                       \\
AVTrack-DeiT \cite{lilearning2024}            & ICML 24                 & 75.3                                      & 68.2                                      & 56.7                                      & 61.5                                      & 55.6                                      & 45.5                                      & 58.6                                      & 59.2                                      & 47.6                                      & \textbf{212}                      & 0.97-1.9                  & 3.5-7.9                     \\
TDA-Track \cite{fu2024prompt}               & IROS 24                 & 75.5                                      & 53.3                                      & 51.4                                      & 61.7                                      & 53.5                                      & 42.3                                      & 49.5                                      & 49.9                                      & 36.9                                      & 114                      & 18.2                      & 9.2                         \\
Sam-DA \cite{fu2024sam}                 & ICARM 24                & 75.3                                      & 64.9                                      & 53.4                                      & 67.3                                      & 59.2                                      & 47.1                                      & 60.4                                      & 59.4                                      & 47.6                                      & 37                       & 27.1                      & 37.4                       \\
SGDViT \cite{yao2023sgdvit}                 & ICRA 23                 & 53.1                                      & 47.2                                      & 38.1                                      & 53.1                                      & 47.9                                      & 37.5                                      & 40.2                                      & 40.6                                      & 32.7                                      & 93                       & 11.3                     & 23.3                      \\
Aba-ViTrack \cite{li2023adaptive}             & ICCV 23                 & 78.4                                      & 72.2                                      & 60.1                                      & 60.4                                      & 57.3                                      & 46.9                                      & 61.3                                      & 63.5                                      & 52.1                                      & 134                      & 2.4                     & 7.9                       \\
HiT-Base \cite{kang2023exploring}                & ICCV 23                 & 62.7                                      & 56.9                                      & 48.2                                      & 49.3                                      & 44.2                                      & 36.4                                      & 48.9                                      & 48.7                                      & 41.1                                      & 156                      & 4.4                    & 42.1                       \\
MAT \cite{zhao2023representation}                     & CVPR 23                 & 80.5 & 75.3           & 61.9 & 64.8                                      & 58.8                                      & 47.7                                      & 57.2                                      & 57.6                                      & 47.1                                      & 56                       & 42.9                    & 88.4                      \\
TCTrack++ \cite{cao2023towards}               & TPAMI 23                & 70.5                                      & 50.8                                      & 46.6                                      & 61.1                                      & 52.8                                      & 41.7                                      & 47.4                                      & 47.4                                      & 37.8                                      & 122                      & 17.6                      & 8.8                         \\
TCTrack \cite{cao2022tctrack}                 & CVPR 22                 & 74.4                                      & 51.2                                      & 47.0                                        & 60.8                                      & 51.9                                      & 40.8                                      & 49.8                                      & 50.0                                      & 37.7                                      & 136                      & 16.9                      & 8.5                         \\
UDAT-BAN \cite{ye2022unsupervised}                & CVPR 22                 & 71.2                                      & 64.9                                      & 51.1                                      & 68.9 & 58.8                                      & 47.2                                      & 61.1                                      & 61.7                                      & 48.4                                      & 41                       & 21.9                     & 54.1                        \\
UDAT-CAR \cite{ye2022unsupervised}                & CVPR 22                 & 68.1                                      & 61.6                                      & 49.6                                      & 68.2                                      & 61.3 & 48.7 & 60.9                                      & 61.3                                      & 48.6                                      & 36                       & 22.3                      & 54.6                        \\
HiFT \cite{cao2021hift}                    & ICCV 21                 & 57.1                                      & 44.5                                      & 40.8                                      & 54.5                                      & 46.7                                      & 37.0                                      & 44.8                                      & 45.2                                      & 35.3                                      & 123                      & 7.2                     & 9.9                       \\
SiamAPN++~\cite{cao2021siamapn++}               & IROS 21                 & 68.9                                      & 57.9                                      & 47.8                                      & 60.2                                      & 51.4                                      & 41.2                                      & 42.7                                      & 41.6                                      & 33.5                                      & 114                      & 8.2                      & 14.7                       \\
SiamCAR \cite{guo2020siamcar}              & CVPR 20                 & 68.7	&62.6	&51.2	&65.8	&59.5	&45.7	&65.8	&65.7	  & 52.3                                      & 37                      & 59.3	&51.3                      \\
Ocean \cite{zhang2020ocean}              & ECCV 20                 & 67.6	&50.3	&44.0	&58.1	&49.9	&38.6	&60.1	&58.9	&47.3                                      & 43	&23.7	&25.8                      \\
\bottomrule
\end{tabular}}
\vspace{-15pt}
\end{table*}

\subsection{Vision Mamba for Tracking}

Given the template image $Z$ and search image $X$, we first embed and flatten them into one-dimensional tokens by a trainable linear projection layer.
This process is called patch embedding and results in $\mathcal{K}$ tokens, formulated by:
\begin{equation}
\begin{aligned}
t^0_{1:\mathcal{K}}&=\mathcal{E} (Z, X) \in \mathbb{R}^{\mathcal{K} \times E}
\end{aligned}
\end{equation}
where $E$ is the embedding dimension of each token.
After obtaining the input tokens $t^0_{1:\mathcal{K}}$, we feed them into the encoding layer, where they are processed through stacked \textit{L} layers of bidirectional Vision Mamba (Vim) encoders.
Let $E^l$ denote the \textit{Vim} layer at layer $l$, where forward propagation procedure involves all tokens from the layer $(l-1)$ via $t_{1:\mathcal{K}}^l=E^l(t_{1:\mathcal{K}}^{l-1}) + t_{1:\mathcal{K}}^{l-1}$.
The detailed structure of the bidirectional Vision Mamba encoders $E^l$ is illustrated on the right side of Fig. \ref{fig_MambaNUT_overview}.
The input $t^0_{1:\mathcal{K}}$ is first normalized and then processed separately through two distinct linear projection layers to obtain the intermediate features $\bm{V}$ and $\bm{Q}$:
\begin{equation}
\begin{aligned}
\bm{V} = Linear^{v}(Norm(t^0_{1:\mathcal{K}})),\\
\bm{Q} = Linear^{q}(Norm(t^0_{1:\mathcal{K}})),\\
\end{aligned}
\end{equation}
Next, we process $\bm{V}$ in both forward and backward directions. In each direction, a 1D convolution followed by a SiLU activation function is applied to $\bm{V}$ to produce $\bm{V}^{\prime}$:
\begin{equation}
\begin{aligned}
\bm{V}_o &= SSM(SiLU(Conv1d(\bm{V}))), \\
\bm{V}_{o}^{\prime} &= \bm{V}_o \odot SiLU(\bm{V}), \\
\bm{Y} &= Linear(\bm{V}_{forward}^{\prime}) + Linear(\bm{V}_{backward}^{\prime}),
\end{aligned}
\end{equation}
where the subscript $o$ are the two scan orientations: forward and backward.
Bidirectional scanning enables mutual interactions among all elements within the sequence, thereby establishing a global and unconstrained receptive field.
The information flow of SSM is described in Eq. \ref{eq_ssm}.
Subsequently, the search region vectors from the output of the last encoder $E^l$ are added element-wise and fed into the tracking head to generate the final tracking results.


\subsection{Tracking head and loss function}

In line with OSTrack \cite{ye2022joint}, we implement a center-based head comprised of multiple Conv-BN-ReLU layers to directly estimate the target's bounding box.
The head outputs local offsets to correct for discretization errors caused by resolution reduction, normalized bounding box sizes, and an object classification score map.
The position with the highest classification score is selected as the object's location, resulting in the final bounding box for the object.

During training, we adopt the weighted focal loss \cite{law2018cornernet} for classification, a combination of $L_1$ loss and Generalized Intersection over Union (GIoU) loss \cite{rezatofighi2019generalized} for bounding box regression.
The total loss function is defined as follows:
\begin{equation}
    \label{eq_loss}
    \mathcal{L}_{total} =  \mathcal{L}_{cls} + \lambda_{iou}\mathcal{L}_{iou} + \lambda_{L_{1}}\mathcal{L}_{L_{1}} + \gamma\mathcal{L}_{ADW}
\end{equation}
where the trade-off parameters are set as $\lambda_{iou}$ = 2 and $\lambda_{L_{1}}$= 5, and  $\gamma$ = 0.00001 in our experiments.

\section{Experiment}

In this section, we provide a thorough evaluation of our method using three nighttime UAV tracking benchmarks: NAT2021~\cite{ye2022unsupervised}, NAT2024-1~\cite{fu2024prompt}, and UAVDark135~\cite{li2022all}.
Our evaluation is performed on a PC that was equipped with an i9-10850K processor (3.6GHz), 16GB of RAM, and an NVIDIA TitanX GPU.
We evaluate our approach by comparing it with 16 state-of-the-art (SOTA) trackers, as detailed in Table \ref{table:overall_performance}.

\subsection{Implementation Details}

\textbf{Model.}
We use Vision Mamba, i.e., Mamba\textsuperscript{\scriptsize\circledR}-Small~\cite{wang2024mamba}, as the backbone to build our MambaNUT tracker for evaluation. The head of MambaNUT consists of a stack of four Conv-BN-Relu layers, with the search region and template sizes set to 256 × 256 and 128 × 128, respectively.

\textbf{Training.}
We use training splits from multiple datasets, including four daytime datasets: GOT-10k \cite{huang2019got}, LaSOT \cite{fan2019lasot}, COCO \cite{lin2014microsoft}, and TrackingNet \cite{muller2018trackingnet}, and three nighttime datasets: BDD100K-Night, SHIFT-Night, and ExDark \cite{loh2019getting}.
Notably, we select the images labeled as "night" from the BDD100K \cite{yu2020bdd100k} and SHIFT \cite{sun2022shift} datasets to construct the BDD100K-Night and SHIFT-Night.
The batch size is consistently set to 32. 
We use the AdamW optimizer with a weight decay of $10^{-4}$, and an initial learning rate of $4\times 10^{-5}$. 
The total number of training epochs is fixed at 300, with 60,000 image pairs processed per epoch. The learning rate is reduced by a factor of 10 after 240 epochs.

\textbf{Inference.}
In the inference phase, following standard practices \cite{zhang2020ocean}, we apply Hanning window penalties during inference to incorporate positional priors into the tracking process. Specifically, we multiply the classification map by a Hanning window of the same size, and the bounding box with the highest score is then selected as the tracking result.

\begin{figure}[!h]
\centering
\vspace{-5pt}
\includegraphics[width=0.475\textwidth]{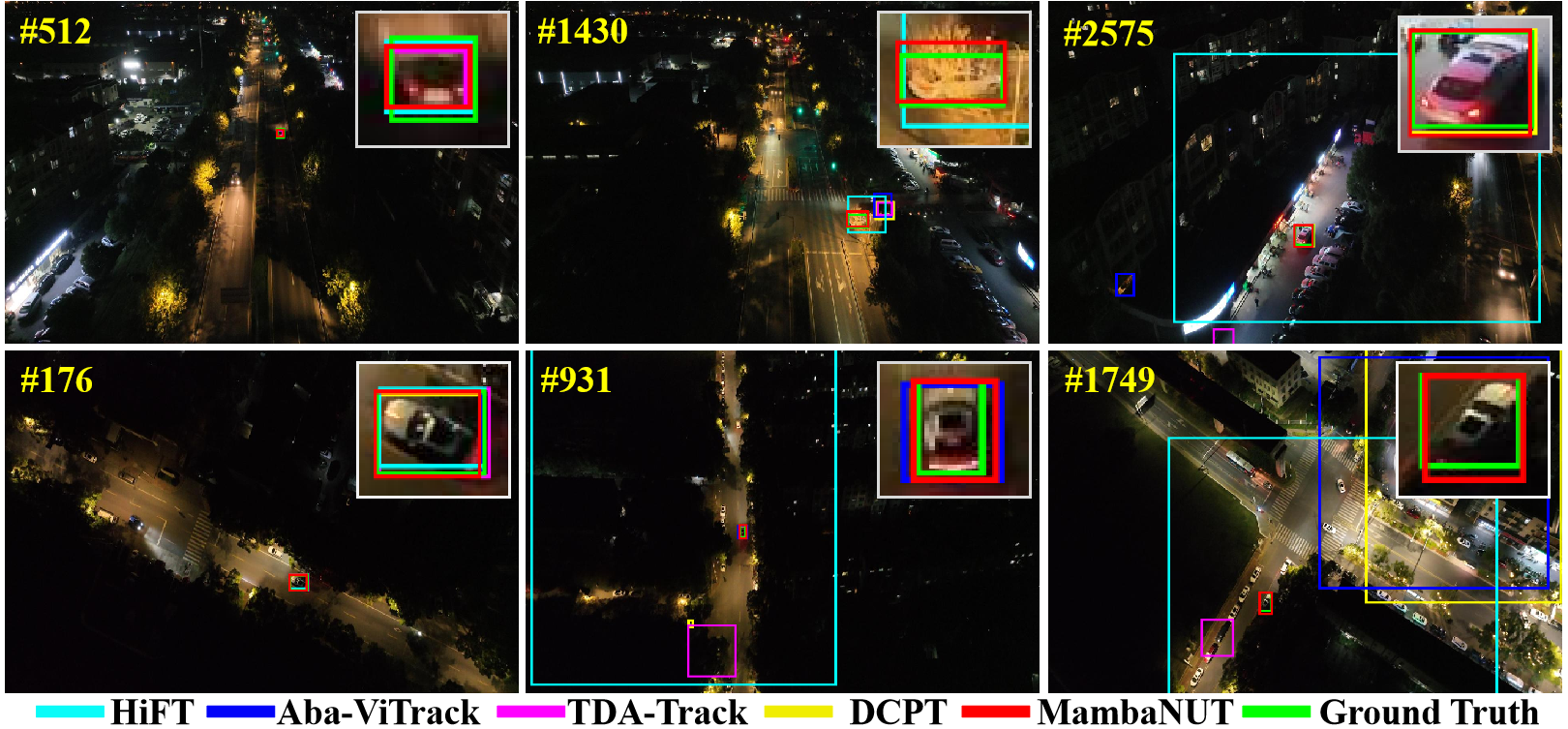}
\caption{Qualitative evaluation on two video sequences from NAT2024-1: L50001 and L50011.}
\label{fig_bbox_vis}
\vspace{-15pt}
\end{figure}

\subsection{Overall Performance}

\textbf{NAT2024-1:}
NAT2024-1 \cite{fu2024prompt} is a long-term tracking benchmark featuring multiple challenging attributes, comprising 40 long-term image sequences with a total of over 70K frames.
As shown in Table \ref{table:overall_performance},
our MambaNUT tracker outperforms 16 state-of-the-art trackers on this benchmark, achieving a precision of 83.3\%, a normalized precision of 76.9\%, and a success rate of 63.6\%, resulting in improvements of 2.4\%, 1.5\%, and 1.5\% over the second-best tracker on these three metrics, respectively.
We also select two representative video sequences from NAT2024-1 for visualization in Fig. \ref{fig_bbox_vis}. As shown, MambaNUT tracks the target objects more accurately than the 4 SOTA trackers.

\textbf{NAT2021:}
NAT2021 \cite{ye2022unsupervised} includes 180 testing videos, offering a challenging and large-scale benchmark for nighttime tracking.
As shown in Table \ref{table:overall_performance}, MambaNUT demonstrates competitive performance compared to the SOTA trackers. It achieves the highest precision (70.1\%) and normalized precision (64.6\%), outperforming the previous top-performing tracker (i.e., DCPT) by more than 1.0\% in both metrics, with only a slight 0.2\% gap in success rate.

\textbf{UAVDark135:} UAVDark135\cite{li2022all} benchmark consists of 135 test sequences and is widely used as the benchmark for nighttime tracking.
From Table \ref{table:overall_performance}, MambaNUT achieved a new SOTA score of 70.0\% in precision and 57.1\% in success rate, with a slight gap in success rate compared to DCPT.

\subsection{Efficiency Comparison}

In Table \ref{table:overall_performance}, we also compare our MambaNUT with SOTA trackers in terms of GPU inference speed, floating-point operations per second (FLOPs), and the number of parameters (Params.) to highlight its superior trade-off between performance and efficiency.
Notably, as AVTrack-DeiT features adaptive architectures, its FLOPs and Params. vary within a range, spanning from the minimum to the maximum values.
As observed, although DCPT achieves comparable performance to our MambaNUT, our method runs in real-time at over 75 fps, more than twice the speed of DCPT, and requires only 1.1 GMac FLOPs and 4.1 million parameters, significantly lower than DCPT's 42 GMacs and 99 million.
While trackers like AVTrack-DeiT and Aba-ViTrack achieve higher tracking speeds than our method, their performance across multiple nighttime UAV tracking benchmarks is significantly inferior.
This comparison in terms of computational complexity also underscores the efficiency of our methods.

\begin{figure}[h]
\centering
\vspace{-5pt}
\begin{subfigure}{0.235\textwidth}
\includegraphics[width=\linewidth]{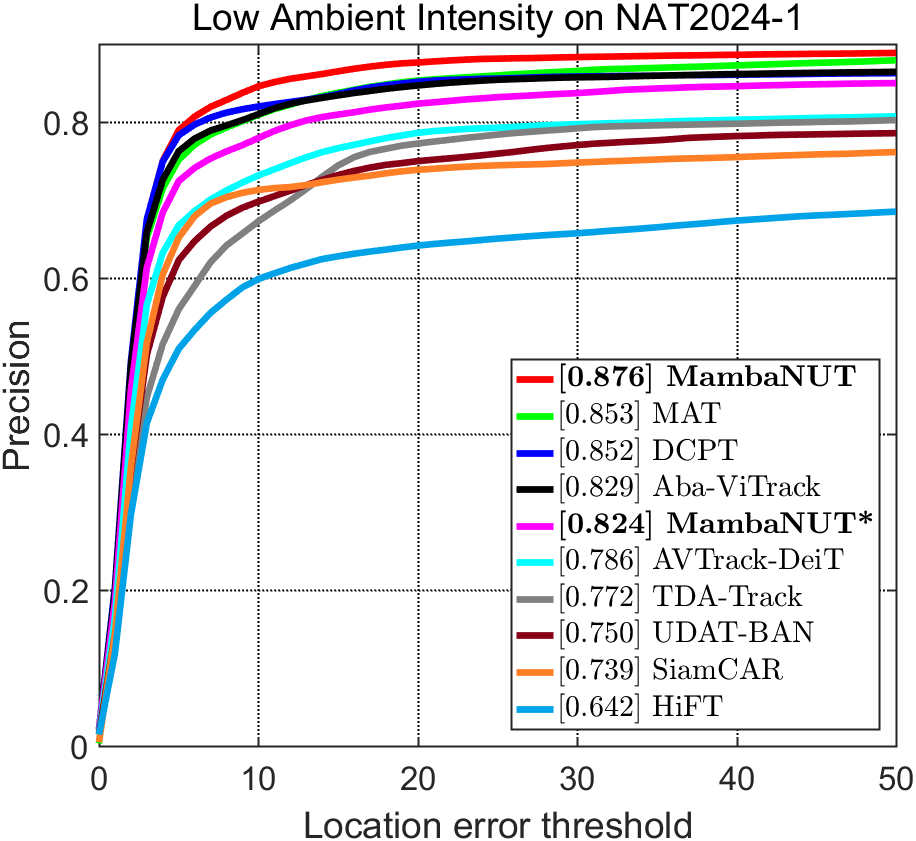} 
    \end{subfigure}
    \begin{subfigure}{0.235\textwidth}
\includegraphics[width=\linewidth]{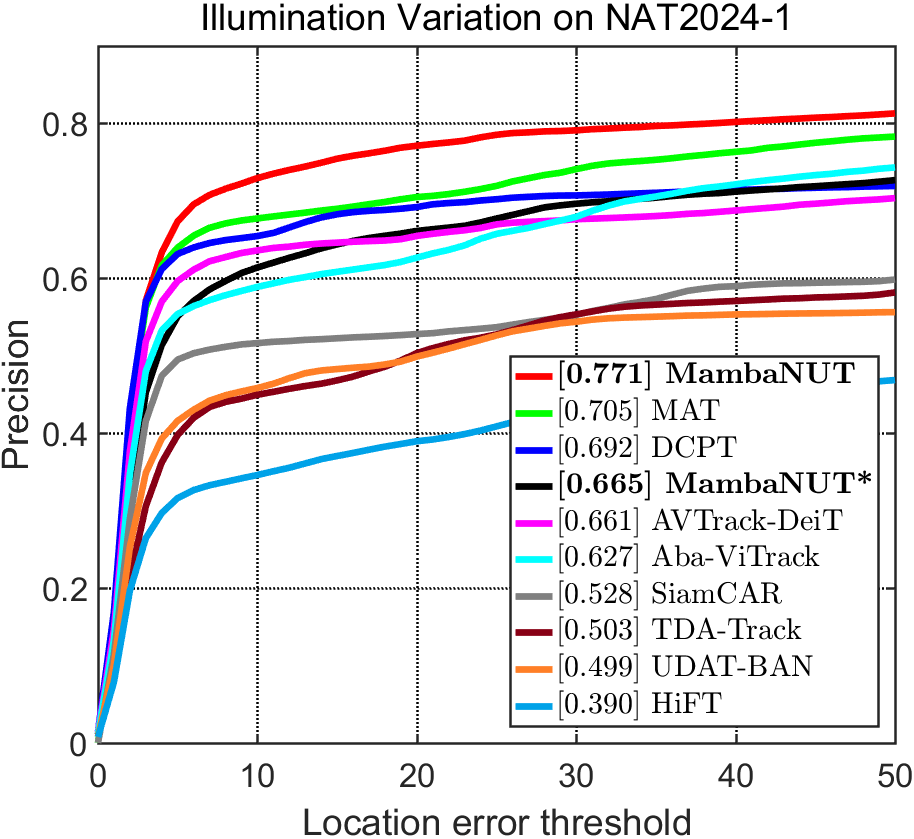} 
\end{subfigure}
\caption{Illumination-oriented evaluation comparison with the 8 SOTA trackers, evaluated on NAT2024-1\cite{fu2024prompt}.}
\label{fig:attrEva}
\vspace{-15pt}
\end{figure}

\subsection{Illumination-Oriented Evaluation}

To further evaluate the performance of MambaNUT in nighttime scenarios, we conduct an analysis focused on the challenges of Low Ambient Illumination (LAI) and Illumination Variation (IV) on NAT2024-1.
Note that we also evaluated MambaNUT without the proposed ACL strategy, referred to as MambaNUT* for comparison.
The precision plots are shown in Fig. \ref{fig:attrEva}.
As observed, our tracker significantly outperforms SOTA trackers in these two attributes, achieving impressive 2.3\% and 6.6\% improvements in precision on LAI and IV, respectively, compared to the second-best tracker.
Notably, incorporating the proposed ACL method leads to substantial improvements of 5.2\% and 10.6\% in precision for LAI and IV, respectively, over MambaNUT*, underscoring its effectiveness.



\subsection{Ablation Study}

\begin{figure}[h]
\centering
\includegraphics[width=0.475\textwidth]{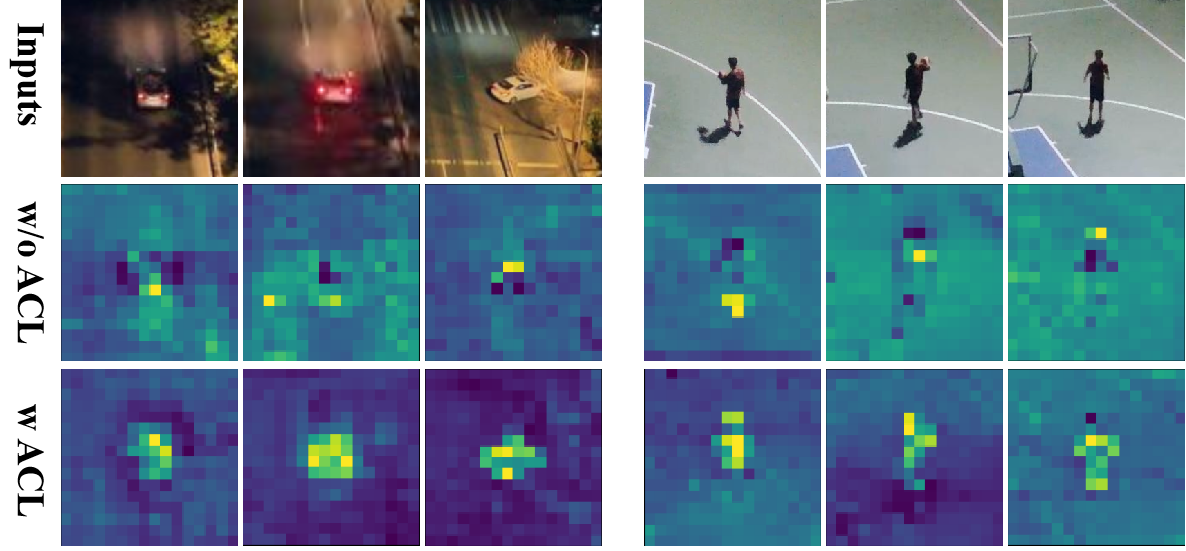}
\caption{Visualization of feature maps generated by MambaNUT without (middle) and with (bottom) the proposed ACL, with the first row displaying input search images from two sequences.}
\label{fig_feat}
\vspace{-7pt}
\end{figure}

\begin{table}[h]
\scriptsize
\centering
\setlength\tabcolsep{9.5pt}
\caption{Impact of Fine-Tuning (FT), Sampling Scheduler (SS), and Loss Scheduler (LS) on baseline tracker performance on NAT2024-1.}\label{ablation_Tab1}
\begin{tabular}{cccc}
\toprule[1pt]
Method                  & Prec. & Norm.Prec & Succ. \\ \hline
baseline & 79.6   & 73.8       & 60.6  \\
baseline+FT & 80.3$_{\uparrow 0.7}$   & 74.6$_{\uparrow 0.8}$       & 61.2$_{\uparrow 0.6}$  \\
baseline+FT+LS & 81.3$_{\uparrow 1.7}$   & 75.1$_{\uparrow 1.3}$       & 61.9$_{\uparrow 1.3}$  \\
baseline+SS& 81.5$_{\uparrow 1.9}$   & 75.3$_{\uparrow 1.5}$       & 61.8$_{\uparrow 1.2}$  \\
baseline+SS+LS& \textbf{83.3}$_{\uparrow 3.7}$   & \textbf{76.9}$_{\uparrow 3.1}$       & \textbf{63.6}$_{\uparrow 3.0}$ \\ \bottomrule[1pt]
\end{tabular}
\vspace{-10pt}
\end{table}

\textbf{Impact of Adaptive Curriculum Learning (ACL) strategy:}
To validate the effectiveness of the proposed adaptive curriculum learning strategy, Table \ref{ablation_Tab1} presents the evaluation results on NAT2024-1, progressively incorporating two levels of curriculum schedulers, i.e., sampling scheduler (SS) and loss scheduler (LS), into the baseline.
Notably, we make additional attempts by first training the model on daytime data and then fine-tuning (FT) it on nighttime data for 50 epochs.
As observed, while FT on the daytime foundation tracker enhances performance, it only achieves results comparable to those with our SS. With the additional application of LS, the improvements become even more significant, with all three metrics gains exceeding 3.0\%.
Fig. \ref{fig_feat} also demonstrates that by incorporating our ACL into the baseline tracker, more robust and discriminative feature representations are achieved, particularly enhancing the consistency of feature distribution across consecutive frames in long-term tracking.
This comparison further demonstrates the effectiveness of our method in enhancing robust feature representations learning under low-light conditions using Mamba.

\begin{table}[h]
\scriptsize
\centering
\setlength\tabcolsep{7.5pt}
\caption{Impact of different loss function schedulers on the performance of MambaNUT.}\label{ablation_Tab2}
\begin{tabular}{ccccc}
\toprule
Method                          & $\mathcal{L}$    & Prec.         & Norm.Prec     & Succ.         \\ \hline
\multirow{4}{*}{MambaNUT} & -     & 79.6          & 73.8          & 60.6          \\
                                & Focal\cite{lin2017focal}  & 80.7$_{\uparrow 1.1}$          & 74.4$_{\uparrow 0.6}$          & 61.3$_{\uparrow 0.7}$          \\
                                & WCE\cite{sudre2017generalised}  & 81.8$_{\uparrow 2.2}$          & 75.5$_{\uparrow 1.7}$          & 61.9$_{\uparrow 1.3}$          \\
                                & \textbf{Ours} & \textbf{83.3}$_{\uparrow 3.7}$   & \textbf{76.9}$_{\uparrow 3.1}$       & \textbf{63.6}$_{\uparrow 3.0}$ \\ \bottomrule[1pt]
\end{tabular}
\end{table}

\textbf{Impact of Loss Function Scheduler:}
To demonstrate the superiority of the proposed ADW loss in performance, we train separately MambaNUT using Focal\cite{lin2017focal} and WCE \cite{sudre2017generalised} loss for comparison.
The evaluation results on NAT2024-1 are shown in Table \ref{ablation_Tab2}.
From table, while using Focal and WCE loss as the loss scheduler improves performance, the best precision improvement is only 2.2\%, and the improvements in norm.precision and success rate remain below 2.0\%, which is far behind our approach, where all three metrics show improvements above 3.0\%.



\begin{figure}[h]
\centering
\includegraphics[width=0.475\textwidth]{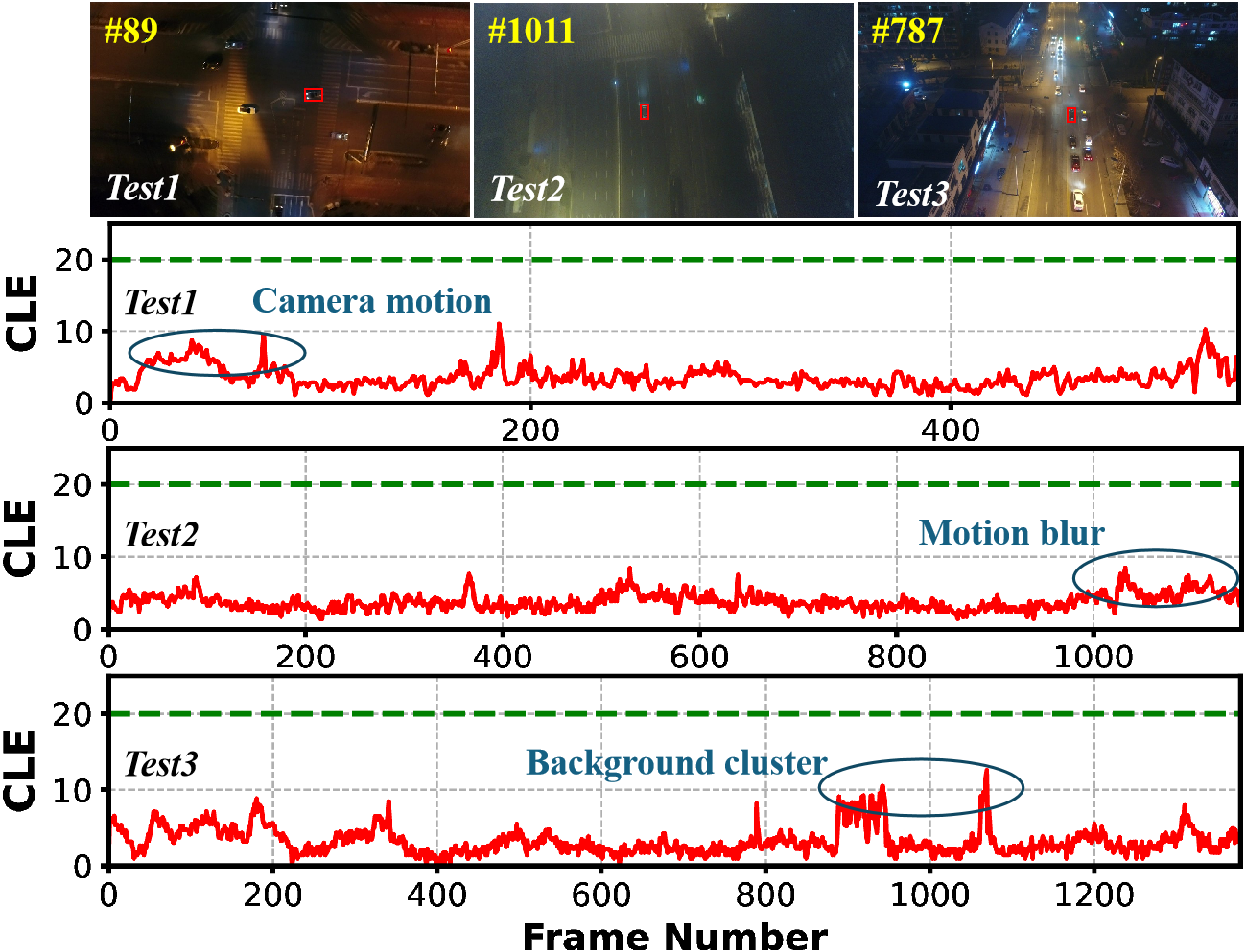}
\caption{Real-world test on an embedded device. Frame-wise performance is illustrated using CLE plots, with errors below the green dashed line (CLE = 20 pixels) considered successful tracking results.}
\label{fig:real_world}
\vspace{-15pt}
\end{figure}

\section{Real-world Tests}
As shown in Fig. \ref{fig:real_world}, we conducted real-world testing by deploying MambaNUT on a standard UAV platform equipped with a NVIDIA Jetson Orin NX to validate its performance.
As shown, the main challenges include partial occlusion, camera motion, and background cluster. Nevertheless, MambaNUT exhibits robust performance, with all test frames maintaining a CLE below 20 pixels.
Additionally, MambaNUT runs in real-time over 30 frames per second.
Real-world testing results demonstrate that MambaNUT is well-suited for edge deployment on UAV platforms, achieving robust tracking performance in complex nighttime circumstances.

\section{Conclusion}

In this work, we propose MambaNUT, a novel Mamba-based nighttime UAV tracking framework that exploits Mamba's exceptional ability to model long-range dependencies with linear complexity.
Additionally, we integrate an adaptive curriculum learning (ACL) strategy into the framework with two schedulers for sampling and loss backward propagation. The proposed scheduler guides the model from imbalance to balance and from easy to hard across daytime and nighttime data. The Adaptive Data Weighted (ADW) loss scheduler employs a weighting scheme based on the size of training data and the IoU of individual instances.
Extensive experiments demonstrate that our MambaNUT achieves state-of-the-art results on three nighttime UAV tracking benchmarks, while offering advantages in computational complexity.

\bibliographystyle{IEEEtran}
\bibliography{IEEEabrv,ref}

\end{document}